# Improved Weighted Random Forest for Classification Problems

Mohsen Shahhosseini[1], Guiping Hu[1]

[1] Department of Industrial and Manufacturing Systems Engineering, Iowa State University, Ames, Iowa, USA

mohsen@iastate.edu

**Abstract.** Several studies have shown that combining machine learning models in an appropriate way will introduce improvements in the individual predictions made by the base models. The key in making well-performing ensemble model is in the diversity of the base models. Of the most common solutions for introducing diversity into the decision trees are bagging and random forest. Bagging enhances the diversity by sampling with replacement and generating many training data sets, while random forest adds selecting random number of features as well. This has made random forest a winning candidate for many machine learning applications. However, assuming equal weights for all base decision trees does not seem reasonable as the randomization of sampling and input feature selection may lead to different levels of decision-making abilities across base decision trees. Therefore, we propose several algorithms that intend to modify the weighting strategy of regular random forest and consequently make better predictions. The designed weighting frameworks include optimal weighted random forest based on accuracy, optimal weighted random forest based on area under the curve (AUC), performance-based weighted random forest, and several stacking-based weighted random forest models. The numerical results show that the proposed models are able to introduce significant improvements compared to regular random forest.

# 1    Introduction

Ensemble learning is defined as a method to combine predictions of multiple machine learning models in order to design a "committee" of decision makers. Intuitively, a committee of decision makers should have lower overall prediction error. This has been investigated in many studies and the results show that improvements can be made on the machine learning models by combining them in an appropriate way [1]. There are several methods to combine base learners including linear/product combination of the learners [2], bagging [3], random forests [4], boosting [5] etc.

It should be noted that including diverse decision makers in the "committee" of base models is necessary to make better judgments, as there will be no improvement from a set of identical models. This is identified as "diversity" of the base learners. It has been proved that well-performing ensemble models demonstrate diversity in base learners as well as achieve superior performance individually. There have been a handful of methods to introduce diversity in the ensemble models including but not limited to bagging, random forests, or boosting methods.

In the bagging method proposed in [3], N samples with replacement of the training data are considered and N training data sets are generated. A learning algorithm, traditionally a decision tree, is built on each of the N training data sets and the final prediction is formed by the simple average or voting over the class label. In essence, bagging tries to incorporate diversity in the ensemble model by creating random differences between the input data sets. On the other hand, random forest learning algorithm [4] adds another source of diversity to the bagging procedure. In the random forest method, in addition to sampling with replacement for creating N training data sets, a random number of features are chosen each time to reduce the amount of correlation between created trees. The final result is again simple average or voting over all the predictions made by the constructed trees of the forest (see Fig. 1).

Although random forest has shown remarkable performance and is used widely in many applications, but it seems that some improvements in the way the base learners are combined can be made to make the random forest predictions even better. Naturally, assuming equal weights for all base learners, i.e. using a simple average for the final predictions, does not seem reasonable. The reason lies behind the randomization of sampling and input feature selection. Specifically, this randomization cannot guarantee that all built trees have the same decision-making abilities [6]. Therefore, having a weighting procedure to weight the trees based on their performance appears reasonable.

**Fig. 1.** Random forest classifier uses majority voting of the predictions made by randomly created decision trees to make the final predictions

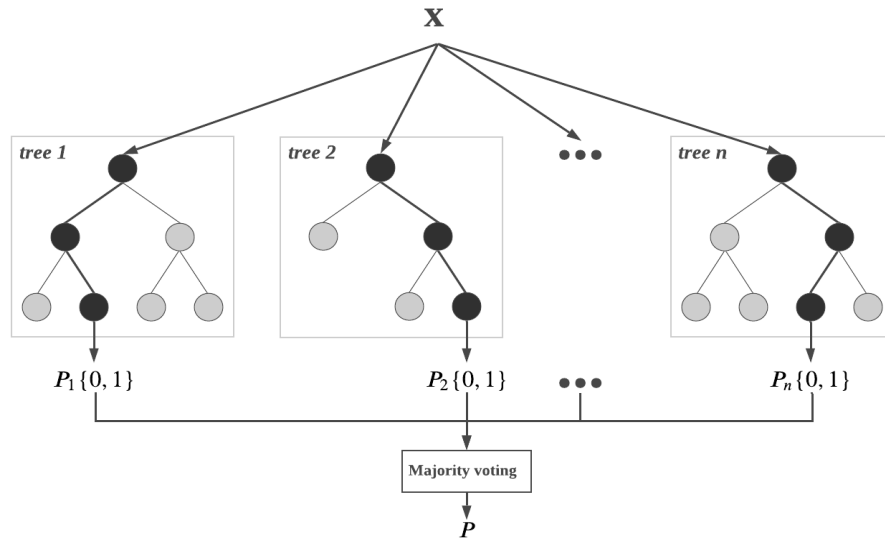

## 2 Background

Several studies have proposed variations of weighted random forest and demonstrated modest improvements in their numerical results compared to traditional random forest predictions [7]. For instance, [8] examined some adjustments in the random forest algorithm to boost its performance. The author modified the evaluation measure and used a combination of five attribute evaluation measures instead of Gini index. In addition, this paper identified instances that are most similar to the target instance and proposed a weighted random forest that used the voting margin of the similar instances as the weights. The numerical results on several data sets showed that the proposed approach is able to improve accuracy and AUC in classification problems.

[6] proposed tree weighted random forest (TWRF) method for classifying high dimensional noisy data. They claimed that since random forest gets affected by noisy data and is prone to make wrong decisions, a new approach for weighting the trees according to their classification ability can be regarded as a remedy. They used out-of-bag (OOB) subset of the training data to evaluate the trees of the forest and eventually used OOB accuracies as the tree weights. The results demonstrated superiority of TWRF compared to regular random forest.

In addition, [9] takes the random forest weighting approaches one step ahead by introducing two weight vectors: a weight vector of classifiers and a weight vector of instances. Specifically, they assigned higher weights to hard-to-classify instances, while assigning performance-based weights to the base tree classifiers. Investigating the method on 28 data sets, it was revealed that the proposed weighted voting method outperforms regular random forest with majority voting, significantly.

In another study, [10] proposed a probabilistic weighted scheme for combining trees of the forest considering four combination methods of majority vote, weighted majority vote, recall combiner and the naïve Bayes combiner. Experimental results with 73 data sets demonstrated that the naïve Bayes combiner is slightly better than other candidates, especially for large balanced data sets, while weighted majority vote is a better weighting method for small unbalanced data sets.

Furthermore, there have been several studies that made use of weighted random forest variations in different applications. [11] proposed a weighted random forest model to identify genetic variants associated with complex disease in high-dimensional data. [12] used performance based weighted random forest model to design an automated trading system that improves the profitability and stability of trading seasonality events. An application of weighted random forest in credit card fraud detection was studied in [13]. Other application areas of weighted random forest models include speech language therapy [14], ischemic stroke lesion segmentation [15], survival analysis [16] and etc.

In this study, we propose optimization, performance, and stacking based weighting mechanisms that try to combine the trees of the forest in a more appropriate manner. The prerequisite for integrating the merits of model combinations is having well-diverse input models. To this end, we try to keep the trees of the forest shallow. This prevents having near identical, i.e. not diverse, trees as input features. The designed models include optimal weighted random forest based on accuracy, optimal weighted random forest based on area under the curve (AUC), performance-based weighted random forest, and several stacking-based weighted random forest models.

## 3  Materials and methods

As stated, to achieve superior performance for the ensemble model, the initial base learners should be diverse. In other words, the correlation between the base models should be low. Therefore, the assumption of the designed ensemble random forest models is that the trees of the forest should be constructed shallow, i.e. they don't acquire large depths. This results in the base decision trees having an acceptable amount of difference from each other. The designed improved weighted random forest models are explained below.

### 3.1  Optimal weighted random forest based on accuracy

The inspiration to design an optimal weighted random forest comes from the optimization model proposed in [17], which minimized mean squared error (MSE) of a linear combination of several base regressors. Here we propose an optimization model to minimize prediction accuracy of a weighted random forest ensemble model for binary classification, in which the weights are the decision variables. The out-of-bag predictions generated by k-fold cross-validation are considered as emulators of unseen test observations and used as inputs to the optimization problem. The mathematical model is as follows.

$$Max\ accuracy\left(Y, \left\lfloor \sum_{j=1}^{k} w_j \hat{Y}_j + 0.5 \right\rfloor \right) \quad (1)$$

$$s.t.$$
$$\sum_{j=1}^{k} w_j = 1,$$
$$w_j \geq 0, \quad \forall j = 1, \dots, k.$$

where $w_j$'s are the weights corresponding to decision tree $j$ ($j = 1, \dots, k$), $Y$ is the vector of actual response values, and $\hat{Y}_j$ is the out-of-bag prediction of decision tree $j$. $accuracy()$ measures the proportion of correct predictions (both true positives and true negatives) among the total number of cases examined. Also, $\left\lfloor \sum_{j=1}^{k} w_j \hat{Y}_j + 0.5 \right\rfloor$ calculates the nearest integer among class labels (0 and 1) for the ensemble model (see Fig. 2).

**Fig. 2.** Optimal weighted random forest classifier uses out-of-bag (OOB) binary predictions of the randomly created decision trees to optimize prediction accuracy

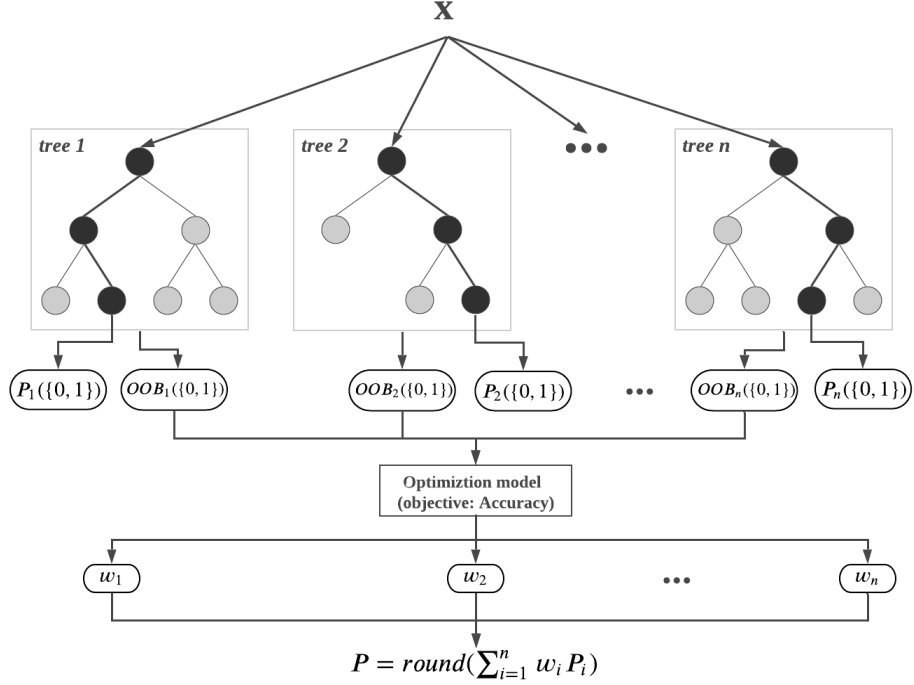

### 3.2 Optimal weighted random forest based on area under the curve (AUC)

Area under the ROC curve (AUC) is a measure used mainly for comparing different classifiers. ROC is a popular curve that plots the trade-off between true positive and false positive rates at different classification thresholds. The area under this graph, i.e. AUC, is useful for comparing binary classifiers as it takes into account all possible thresholds. In addition, the accuracy has an intrinsic drawback of reporting very high accuracy when classifying highly imbalanced data sets [18]. The following optimization model intends to find the optimal weights of combining trees of a random forest model by optimizing the ensemble's AUC.

$$Max \ AUC\left(Y, \sum_{j=1}^{k} w_j \hat{P}_j\right) \quad (2)$$
$$s.t.$$
$$\sum_{j=1}^{k} w_j = 1,$$
$$w_j \geq 0, \quad \forall j = 1, \dots, k.$$

The $\hat{P}_j$ in the above formulation refers to the out-of-bag probability vector of each base classifier, and $AUC()$ calculates the area under the ROC curve for the built ensemble. Fig. 3 depicts the proposed framework to create an optimal weighted random forest using out-of-bag probabilities of true class.

**Fig. 3.** Optimal weighted random forest classifier uses out-of-bag (OOB) probability predictions of true class made by randomly created decision trees to optimize AUC

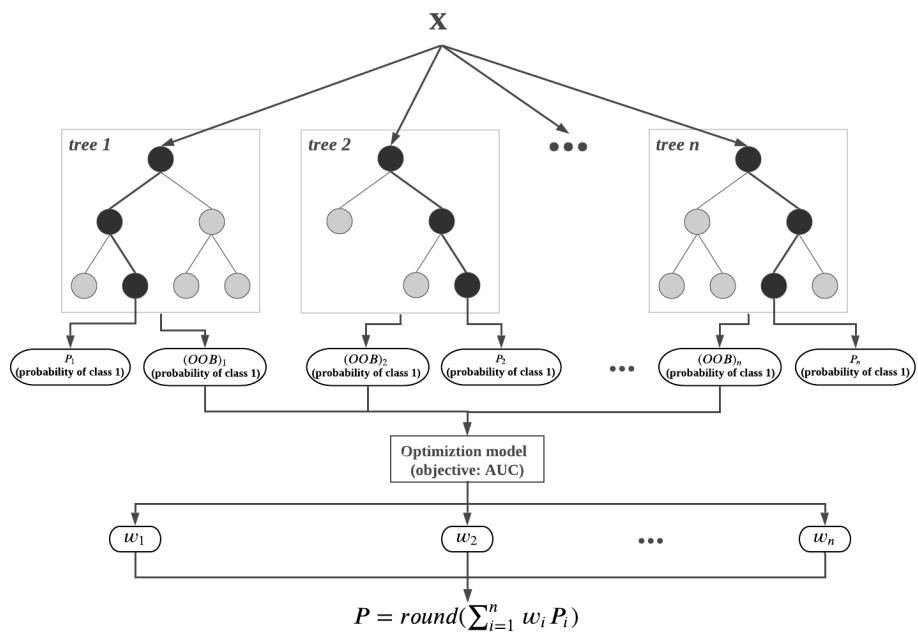

### 3.3 Performance-based weighted random forest model

We design a performance-based weighted random forest models using out-of-bag accuracy. Each classifier's out-of-bag performance is measured with accuracy and the normalized accuracy of each classifier is used as its weight in forming the ensemble. The procedure is demonstrated in Fig. 4.

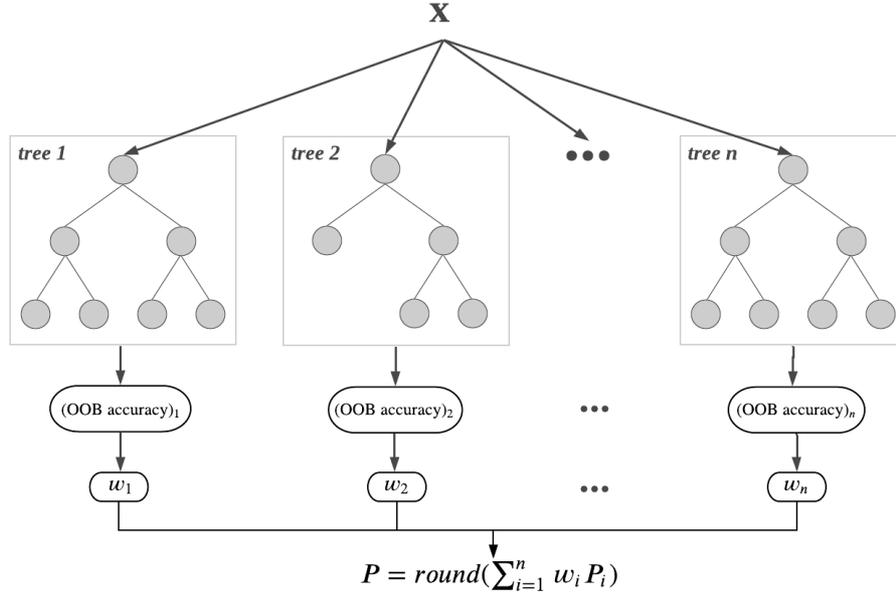

**Fig. 4.** Performance-based weighted random forest classifier uses accuracy of out-of-bag (OOB) predictions as the ensemble weights

### 3.4 Stacking-based random forest models

Stacking is a well-known ensemble learning procedure which combines multiple base learners by performing at least another level of learning task. It uses out-of-bag predictions of the base learners, and the actual response values of the training data to form the independent and dependent variables of the 2nd-level learning task, respectively [19]. Here, we use the following steps to make use of the out-of-bag predictions made by trees of the forest and train another machine learning model on top of them to end up with an improved random forest (see Fig. 5).

1. Learn a random forest model using the training data
2. Obtain decision trees of the forest and generate out-of-bag predictions for each of them by using $k$-fold cross-validation.
3. Create a new data set with out-of-bag predictions as the input variables and actual response values of data points in the training set as the response variable.
4. Learn a second-level machine learning model on the created data set and make predictions for unseen test observations.

**Fig. 5.** Stacking-based weighted random forest classifier trains a second level machine learning model on the out-of-bag (OOB) predictions made by each randomly created decision tree

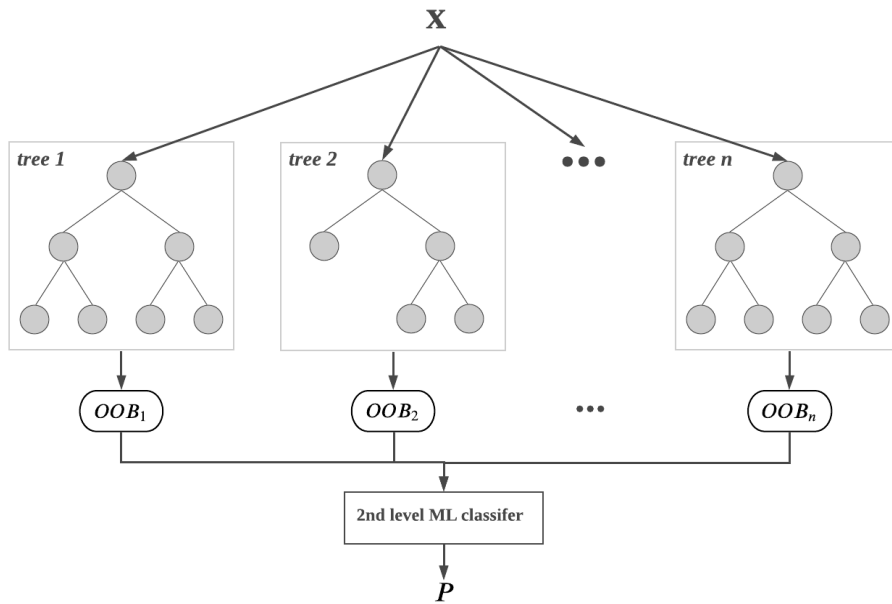

We have selected three machine learning models as the second level classifier which include random forest, logistic regression, and K-nearest neighbors. In addition, two scenarios are considered for each 2nd level classifier: using binary classifications of out-of-bag predictions or using out-of-bag predictions of the probability an observation belongs to the majority class. In the second case instead of having binary predictions as input variables, the probability of the true class (class 1) is used.

## 4    Results and Discussion

To examine the performance of proposed improved weighted random forest classifiers, 25 public binary classification data sets from UCI machine learning repository [20] were used. Minimal pre-processing tasks including treating missing values, one hot encoding etc. were performed on each data set to make them ready for training classification models. To evaluate the true performance of built models, 25% of each data set is reserved as the test set and the remaining 75% is used to build and optimize ensemble models.

For training the regular random forest and subsequently $n$ randomly created decision trees, the number of trees ($n$) is set at 100 trees and for keeping the trees shallow and hence not correlated with each other, the maximum depth of the trees is set at the half of the common choice for the maximum depth of random forest trees, i.e. square root of the number of features ($\sqrt{p} / 2$). Out-of-bag predictions of the decision trees are generated using 5-fold cross-validation. For solving the optimization problems, Sequential Least Squares Programming algorithm (SLSQP) from Python's SciPy optimization library was used [21]. Finally, the entire process is repeated 50 times to avoid biased results due to the randomness in partitioning the data set into training and test sets, and in forming decision trees. The reported numerical results are the average of 50 runs for each data set.

**Table 1.** Details of example data sets downloaded from UCI machine learning repository

| # | Data set | Size | Features | Class 0 (%) | Class 1 (%) |
|---|---|---|---|---|---|
| 1 | Acute inflammations [22] | 120 | 6 | 42% | 58% |
| 2 | Adult | 48842 | 14 | 24% | 76% |
| 3 | APS failure at Scania trucks | 60000 | 171 | 2% | 98% |
| 4 | Audit [23] | 777 | 18 | 39% | 61% |
| 5 | Australian credit approval | 690 | 14 | 44% | 56% |
| 6 | Autism screening adult | 704 | 21 | 30% | 70% |
| 7 | Bank marketing [24] | 45211 | 17 | 12% | 88% |
| 8 | Breast cancer (Wisconsin) | 569 | 31 | 37% | 63% |
| 9 | Breast cancer (Yugoslavia) | 286 | 9 | 29% | 71% |
| 10 | Census-income (KDD) | 99762 | 40 | 6% | 94% |
| 11 | Cervical cancer (risk factors) [25] | 858 | 36 | 9% | 91% |
| 13 | Climate model simulation crashes | 540 | 18 | 9% | 91% |
| 12 | Cylinder bands | 512 | 39 | 42% | 58% |
| 14 | Default of credit card clients [26] | 30000 | 24 | 22% | 78% |
| 15 | Divorce predictors [27] | 170 | 54 | 49% | 51% |
| 16 | Drug consumption (quantified) [28] | 1885 | 32 | 33% | 67% |
| 17 | EEG eye state | 14980 | 15 | 45% | 55% |
| 18 | Electrical grid stability simulated | 10000 | 14 | 36% | 64% |
| 19 | Extension of Z-Alizadeh Sani [29] | 303 | 59 | 29% | 71% |
| 20 | Hepatitis | 155 | 19 | 20% | 80% |
| 21 | Horse colic | 368 | 27 | 36% | 64% |
| 22 | HTRU2 [30] | 17898 | 9 | 9% | 91% |
| 23 | Indian liver patient dataset (ILPD) | 583 | 10 | 28% | 72% |
| 24 | Internet advertisements | 3279 | 1558 | 14% | 86% |
| 25 | Ionosphere | 351 | 34 | 36% | 64% |

Table 1 represents the details of 25 example data sets obtained from UCI machine learning repository including the data set size, number of features and the proportion of class labels. As it can be seen in this table, the chosen data sets cover data sets with different levels of size and class label proportions.

The complete experimental results of all designed ensemble models applied on the example data sets are shown in Table 2. Stacking-based ensembles have created in two ways: using binary OOB predictions and using OOB probability predictions of the true class label. As the results suggest in 22 out of 25 considered data sets at least one of the designed models provide improvements over regular random forest (first column of the table). In addition, it appears that the stacking-based random forest with a second random forest model as the second level classifier when having binary OOB binary predictions outperforms other models more often (10 data sets out).

**Table 2.** Experimental results of designed improved random forest classifiers compared to regular random forest classifier. The best-performing classifier for each data set is highlighted. The last row shows the average accuracy of all models considering all data sets.

| # | RF[1] | Optimal WRF (Acc.)[2] | Optimal WRF (AUC)[3] | RF stacked RF (binary)[4] | RF stacked RF (prob.)[5] | Log. stacked RF (binary)[6] | Log. stacked RF (prob.)[7] | KNN stacked RF (binary)[8] | KNN stacked RF (prob.)[9] | Performance based WRF[10] |
|---|---|---|---|---|---|---|---|---|---|---|
| 1 | 91.00% | 98.20% | 98.20% | 100.00% | 97.00% | 100.00% | 90.27% | 100.00% | 100.00% | 97.73% |
| 2 | 83.75% | 82.12% | 78.67% | 85.18% | 82.73% | 84.67% | 83.27% | 84.28% | 84.75% | 82.12% |
| 3 | 99.21% | 99.21% | 99.20% | 99.24% | 99.21% | 99.21% | 99.20% | 99.23% | 99.20% | 99.21% |
| 4 | 99.69% | 99.69% | 99.67% | 99.94% | 99.95% | 99.95% | 99.95% | 99.95% | 99.95% | 99.70% |
| 5 | 85.86% | 85.69% | 74.42% | 84.88% | 83.87% | 86.24% | 84.84% | 85.11% | 85.57% | 85.99% |
| 6 | 99.92% | 99.62% | 99.62% | 100.00% | 100.00% | 100.00% | 100.00% | 99.91% | 99.93% | 99.62% |
| 7 | 88.92% | 89.25% | 88.43% | 90.04% | 85.44% | 90.02% | 89.25% | 89.74% | 89.78% | 89.25% |
| 8 | 94.69% | 94.73% | 94.22% | 94.77% | 93.93% | 94.71% | 94.76% | 94.83% | 94.69% | 94.76% |
| 9 | 74.89% | 75.43% | 73.00% | 72.89% | 62.49% | 72.54% | 72.43% | 73.26% | 73.77% | 75.31% |
| 10 | 94.79% | 94.72% | 94.60% | 95.39% | 95.00% | 95.53% | 95.33% | 95.06% | 95.19% | 94.71% |
| 11 | 90.49% | 90.47% | 90.31% | 89.51% | 25.13% | 90.14% | 90.12% | 90.22% | 90.08% | 90.47% |
| 12 | 91.04% | 91.04% | 90.79% | 90.39% | 91.07% | 91.32% | 91.01% | 90.89% | 91.41% | 91.04% |
| 13 | 72.01% | 69.90% | 66.67% | 73.67% | 70.06% | 69.78% | 69.01% | 71.42% | 71.93% | 69.82% |
| 14 | 80.25% | 80.69% | 79.07% | 81.21% | 81.39% | 82.14% | 80.76% | 79.69% | 79.82% | 80.69% |
| 15 | 97.26% | 97.26% | 97.26% | 97.02% | 97.12% | 97.35% | 97.40% | 97.26% | 97.26% | 97.26% |
| 16 | 70.19% | 71.18% | 69.28% | 69.76% | 69.36% | 71.22% | 70.17% | 68.48% | 70.32% | 71.29% |
| 17 | 66.29% | 64.23% | 59.89% | 73.97% | 64.50% | 70.70% | 66.22% | 71.46% | 67.68% | 64.31% |
| 18 | 99.97% | 96.10% | 96.23% | 99.97% | 99.97% | 99.98% | 99.97% | 99.98% | 99.98% | 98.52% |
| 19 | 82.55% | 83.47% | 77.24% | 84.79% | 81.71% | 80.11% | 80.29% | 83.61% | 83.84% | 83.76% |
| 20 | 83.33% | 83.11% | 81.50% | 79.56% | 79.83% | 80.11% | 80.56% | 79.83% | 83.00% |
| 21 | 85.47% | 85.23% | 77.36% | 84.83% | 37.09% | 82.53% | 37.33% | 85.63% | 37.09% | 85.25% |
| 22 | 97.07% | 96.78% | 95.66% | 97.77% | 97.72% | 97.77% | 97.61% | 97.63% | 97.72% | 96.85% |
| 23 | 71.74% | 71.75% | 70.99% | 68.10% | 52.06% | 69.52% | 28.15% | 68.54% | 51.06% | 71.77% |
| 24 | 96.88% | 96.93% | 96.79% | 97.50% | 97.35% | 97.26% | 97.27% | 97.27% | 97.37% | 96.93% |
| 25 | 92.86% | 92.77% | 89.66% | 92.61% | 91.59% | 90.32% | 90.09% | 92.30% | 92.00% | 92.75% |
| **Ave.** | **87.61%** | **87.58%** | **85.55%** | **88.12%** | **81.42%** | **87.73%** | **83.39%** | **87.85%** | **85.21%** | **87.68%** |

[1] Regular random forest classifier
[2] Optimal weighted random forest based on accuracy
[3] Optimal weighted random forest based on AUC
[4] Stacking-based random forest with random forest as the 2nd level classifier using binary OOB predictions
[5] Stacking-based random forest with random forest as the 2nd level classifier using probability OOB predictions
[6] Stacking-based random forest with logistic regression as the 2nd level classifier using binary OOB predictions
[7] Stacking-based random forest with logistic regression as the 2nd level classifier using probability OOB predictions
[8] Stacking-based random forest with KNN as the 2nd level classifier using binary OOB predictions
[9] Stacking-based random forest with KNN as the 2nd level classifier using probability OOB predictions
[10] Performance-based weighted random forest

Fig. 6 intends to better compare the performance of proposed improved random forest classifiers with regular random forest by depicting average accuracy of designed models over all data sets. Fig. 6 (a) presents a chart comparing average accuracy scores of all models with regular random forest classifier. Based on this figure it is evident that RF stacked RF using binary OOB predictions outperforms all designed classifiers as well as regular random forest. Fig. 6 (b) compares the average improvements each model could provide over regular random forest classifier. This figure suggests that four of the designed models improve the prediction accuracy of regular random forest with RF stacked RF (binary) as the best-performing model. This classifier could improve regular random forest predictions by 0.5%.

**Fig. 6.** Comparing proposed improved random forest classifiers with regular random forest

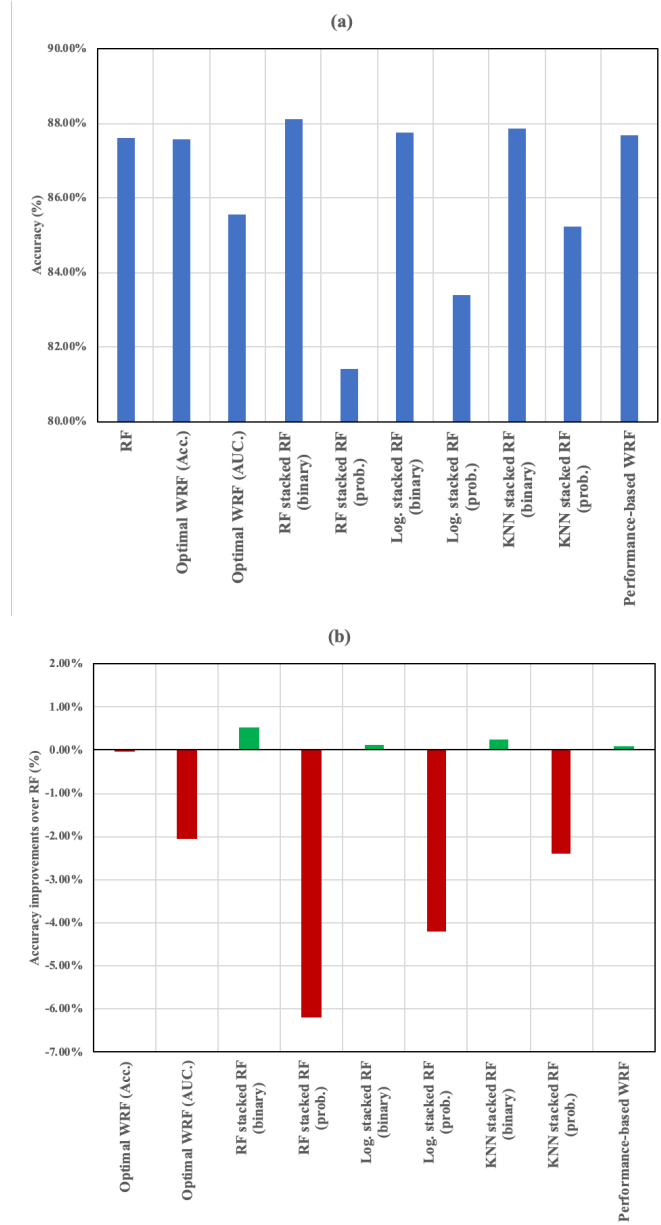

## 5  Conclusion

This study was an attempt to improve well-performing machine learning model, random forest, as a classifier. To this end, several models based on ensemble learning were designed here. The proposed models include optimal weighted random forest using out-of-bag accuracy and AUC, stacking-based random forest, and performance-based random forest. The experimental results of applying the designed models on 25 public data sets demonstrated that four of the designed models could provide some improvements over the regular random forest classifier. From them, a stacking-based random forest model which trains a $2^{nd}$ level random forest on inner randomly created decision trees outperformed all designed models. The possible future research directions following this study could be as follows.

- Using other optimization tools to find a better optimal solution for the weights ([31], [32], [33], [34], [35]);
- Adding instance-weighting to the developed framework to assign higher weights to hard-to-classify instances ([8], [9]);
- Developing a similar framework to improve random forest regressor; and
- Applying similar concept on other fields or research such as data envelopment analysis (DEA) ([36], [37], [38], [39])
- Combining bagged and boosted trees to improve both bias and variance of the predictions.